\documentclass[runningheads]{llncs}
\usepackage{graphicx}
\usepackage{makecell}
\usepackage{hyperref}
\usepackage{amsmath} %for multiple lines in Eq.

\usepackage{ifthen}
\usepackage{xcolor}

\newboolean{showcomments}
\setboolean{showcomments}{false}

\ifthenelse{\boolean{showcomments}}
  {\newcommand{\nb}[3]{
  {\color{#2}\small\fbox{\bfseries\sffamily\scriptsize#1}}
  {\color{#2}\sffamily\small$\triangleright~$\textit{\small #3}$~\triangleleft$}
  }
  }
  {\newcommand{\nb}[3]{}
  }

\newcommand\blfootnote[1]{%
  \begingroup
  \renewcommand\thefootnote{}\footnote{#1}%
  \addtocounter{footnote}{-1}%
  \endgroup
}

\begin{document}
\title{Map-merging Algorithms for Visual SLAM: Feasibility Study and Empirical Evaluation %\thanks{This work was supported by Russian Science Foundation (project \#16-11-0048)}
}
\titlerunning{Map-merging Algorithms for Visual SLAM...}
\author{Andrey Bokovoy\inst{1,2}\orcidID{0000-0002-5788-5765} \and
Kirill Muraviev\inst{1,3}\orcidID{0000-0001-5897-0702} \and
Konstantin Yakovlev\inst{1,3,4}\orcidID{0000-0002-4377-321X}}
\authorrunning{A. Bokovoy et al.}
\institute{Artificial Intelligence Research Institute, Federal Research Center for Computer Science and Control of Russian Academy of Sciences, Moscow, Russia.
\and
Peoples' Friendship University of Russia (RUDN University), Moscow, Russia
\and
Moscow Institute of Physics and Technology, Dolgoprudny, Russia
\and
National Research University Higher School of Economics, Moscow, Russia
\email{\{bokovoy,yakovlev\}@isa.ru, kirill.mouraviev@yandex.ru}}

\maketitle              % typeset the header of the contribution

\begin{abstract}
Simultaneous localization and mapping, especially the one relying solely on video data (vSLAM), is a challenging problem that has been extensively studied in robotics and computer vision. State-of-the-art vSLAM algorithms are capable of constructing accurate-enough maps that enable a mobile robot to autonomously navigate an unknown environment. In this work, we are interested in an important problem related to vSLAM, i.e. map merging, that might appear in various practically important scenarios, e.g. in a multi-robot coverage scenario. This problem asks whether different vSLAM maps can be merged into a consistent single representation. We examine the existing 2D and 3D map-merging algorithms and conduct an extensive empirical evaluation in realistic simulated environment (Habitat). Both qualitative and quantitative comparison is carried out and the obtained results are reported and analyzed.

\end{abstract}
\section{Introduction}

Simultaneous localization and mapping (SLAM)\blfootnote{camera-ready version as submitted to RCAI-2020} is one of the major problem in mobile robotics as the ability to construct a map and localize itself on this map is vital for autonomous navigation in a wide range of scenarios. Different perception capabilities require different SLAM techniques. For example, different approaches exist for LiDAR~\cite{hess2016real,alismail2014continuous}, RGB-D cameras~\cite{hu2012robust,engelhard2011real}, sonars~\cite{ribas2006slam} and other sensors. 

One of the most challenging SLAM formulation is monocular vision-based SLAM (vSLAM) when only the data from a single camera is available. This problem is of particular importance when it comes down to compact mobile robots that could not be equipped with wide array of sensors due to low payload and battery limitations. Often a single camera is the only sensor that a robotic system is equipped with. Indeed this brings its own challenges and issues~\cite{pirchheim2013handling,namdev2013multibody} that need to be taken into account, when designing vSLAM methods.

One of the common problems that is addressed in more advanced applications of vSLAM algorithms, is the problem of combining several local maps into a single global map. This is the case for at least two different scenarios (see~\figurename~\ref{fig:local_maps_issue}): 

\begin{figure}[t]
    \centering
    \includegraphics[width=1.0\textwidth]{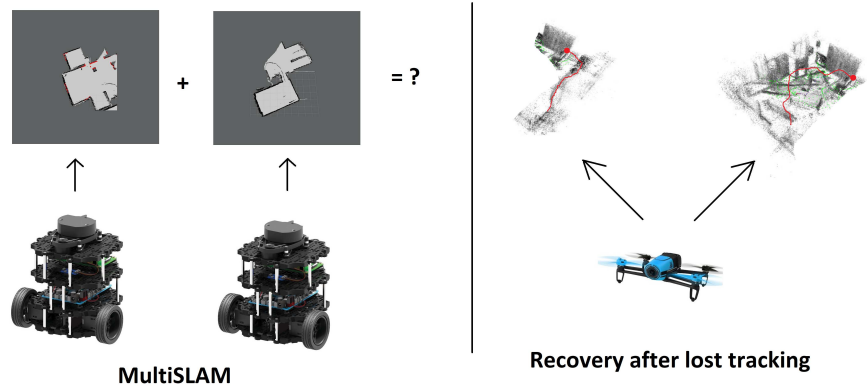}
    \caption{Different scenarios in which the need for map-merging arises. Left: two robots navigate the same environment via different trajectories. Right: a single robot has to recover after the tracking has lost.}
    \label{fig:local_maps_issue}
\end{figure}

\begin{enumerate}
    \item A single robot moves through an unknown environment. The localization and mapping get lost several times (e.g. in low structured visual environment) causing the algorithm to reinitialize each time and build a new local map without taking into account the previously built ones;  
    \item Multiple robots operate in an unknown environment. Each robot builds its own local map without knowing other robots' maps and positions.  
\end{enumerate}

To construct a consistent global map in both scenarios one needs to find overlaps between the local maps, adjust the robot(s)' position(s) appropriately (translate from local coordinate frame to a global one) and construct a single joint map. This problem is known as \textbf{map-merging}~\cite{ozkucur2009cooperative,lee2012probabilistic,dinnissen2012map}. Noteworthy that within a visual SLAM framework, both 2D and 3D maps can be merged.

Indeed there exist methods and algoithms that address map-merging problem. However, most of these algorithms are typically validated on an exact single/multi robot system, whilst quantitative and qualitative comparison with the other approaches is left out of the scope due to the complexity of the experimental evaluation. To this end, we focus on analyzing different map-merging methods (both 2D and 3D) by running a wide set of reproducible experiments in a realistic simulated environment. The contribution of the paper is two-fold. First, we create a dataset that enables the evaluation of map-merging algorithms by using a photorealistic simulated environment Habitat~\cite{habitat19iccv} (and make this dataset available to the community). Second, we conduct a thorough empirical evaluation and comparison of 3 different map-merging algorithms paired with 2 different vSLAM methods on the created dataset. Both qualitative and qualitative comparison are carried out and reported.

This paper is organized as follows: Section \ref{text:related_work} describes the current state of the research in general vSLAM area and map-merging in particular. Section \ref{text:problem_statement} states the problem formally. Section \ref{text:methods_overview} describes the methods that were evaluated as well as the metrics used. Section \ref{text:experiments} presents the experimental setup and the results of the empirical evaluation as well as our interpretation and analysis. Section \ref{text:conclusion} concludes.

%, as well as provides an information about realistic simulated environments.

\section{Related Work}\label{text:related_work}

\subsection{vSLAM Algorithms}

Vision-based simultaneous localization and mapping algorithms vary depending on the vision-sensor which is used to gather the data, differentiating between the monocular, stereo and RGB-D settings. In each case an input for a map-merging algorithm is different, thus the vSLAM pipeline is different as well.

%\begin{figure}[t]
%    \centering
%    \includegraphics[width=1.0\textwidth]{img/slam_ex.eps}
%    \caption{Examples of different types of vSLAM. a) Monoclular feature-based SLAM, b) LiDAR %SLAM, c) RGB-D SLAM }
%    \label{fig:vslam_examples}
%\end{figure}{}

vSLAM methods that rely on a videoflow from a single camera (monocular vSLAM) can be divided into two major groups: feature-based and dense. Feature based methods, e.g.~\cite{mur2017orb,klein2007parallel}, are fast and accurate enough in terms of localization. However the resultant maps are sparse and lack details. On the other hand, dense methods, like~\cite{engel2014lsd,engel2017direct}, provide highly detailed maps, at the cost of higher runtimes. Thus their application for real-time on-board processing is limited when it comes to compact robotic systems that are not equipped with powerful computers. 

In stereo-based SLAM~\cite{engel2015large,gomez2019pl} a disparity map from a pair of images is obtained and further used for computing a depth map. This map is then used for localization and mapping purposes. Naturally, processing two video streams instead of one adds additional computation costs.

The RGB-D cameras~\cite{zhang2012microsoft,keselman2017intel} construct depth maps on-the-fly and provide them to a vSLAM algorithm along with a single RGB video stream. As a result, RGB-D vSLAM methods~\cite{kerl15iccv,whelan2015real} are accurate in terms of localization, provide detailed maps and do not incur extra computational costs. However, RGB-D sensors are notably larger and heavier than RGB cameras and consume more power, which limits their usage onboard of compact mobile robots. 

Overall, when it comes to hard constraints imposed on robot's size, weight and power the most reasonable choice is \emph{i}) equipping it with a single compact and light-weight RGB sensor, i.e. a video-camera; \emph{ii}) relying on monocular vSLAM algorithms. One of the way to enhance the performance of the latter is to augment them with modern depth reconstruction algorithms that are fast enough to be executed on-board in real time. Such algorithms, indeed, exists -- see ~\cite{wofk2019fastdepth,bokovoy2019real} and they typically rely on using convolutional neural networks (CNN). Within this approach an image from monocular camera is combined with a CNN reconstructed depth map, which turns monocular camera into an RGB-D sensor. This provides a reasonalbe trade-off between the localization/map accuracy, map density and processing speed. In this work we chose this approach as a baseline for further evaluation of map-merging algorithms.

%Noteworthy modern on-board computers allows the usage of graphical purpose unit (GPU) accelerated machine learning algorithms for computer vision tasks. Despite it's computation power, such computers are small (credit card sized) and have low power consumption, which makes them suitable for small mobile robotics usage. That leads to the growing interest in using convolutional neural networks (CNN) for single image depth reconstruction for further usage in real-time vSLAM algorithms~\cite{wofk2019fastdepth,bokovoy2019real}. Within this approach an image from monocular camera is combined with a CNN reconstructed depth map, which turns monocular camera into an RGB-D one. We found, that this approach keeps trade-off between accuracy, maps' density and processing speed, so we choose this approach as a baseline for map-merging algorithm evaluation. 

\subsection{Map-merging Algorithms}

Generally the input of a map-merging algorithm is a set of maps constructed by a vSLAM method(s). These maps are typically represented as 3D pointclouds~\cite{rusu20113d}. Meanwhile, in most cases, map-merging algorithms operate with occupancy grids~\cite{konolige1997improved}, which are obtained by selecting a plane, e.g. a ground plane in case of a wheeled robot, slicing the point cloud and, finally, forming a grid representation of the slice. Each element of such grid corresponds to a probability for particular cell being an obstacle, free space or unknown environment. This grid is usually represented as an image and computer vision techniques are used in order to analyse its structure.

In~\cite{birk2006merging} the problem of multi-robot mapping is considered when the integration of different maps obtained by different robots into a joint one is needed. The authors propose a method to estimate map similarity as well as a stochastic search algorithm (Carpin's Adaptive Random Walk) for merging. This algorithm finds the maximum overlap between the input maps and calculates the transform between them in order to combine the robtos' poses and maps into a single representation model. The method is evaluated on 6 real robots. The algorithm is computationally expensive and its success rate was low when some of the input parameters (threshold for random walk) was set inappropriately. 

In~\cite{saeedi2015occupancy} the occupancy maps are considered as images and each map is processed through 3 major steps: preprocessing (edge detection, edge smoothing), finding overlaps (segmentation, segment verification, cross validation) and relative pose finding. The edge detection is done with Canny edge detector. Detected edges are are then segmented into blocks, which are compared across the maps in order to find the overlaps. If an overlap is found, the relative pose is calculated using the translation and rotation of one of the maps. The algorithm is fast but not robust as it relies on Canny detector which is known to perform poorly in un-structured environments. 

The robust overlap estimation is addressed in~\cite{Horner2016}. The map comparison is done with the ORB feature detector~\cite{rublee2011orb} and the RANSAC algorithm is used in order to find transformation, merge the maps and find the pose transformation. The code of this method is open-sourced.

The robustness of the overlap procedure can be increased by considering 3D maps, which contain much more features compared to their 2D slices. Indeed, this comes at a cost of higher runtimes. In~\cite{article} the poincloud maps are represented as pose graphs and the overlaps are found by per-node comparison of two maps, where each node contains local sub-maps. When the match is found 2 maps are merged into a single graph and the global optimization is carried out. Both these operations are computationally expensive, thus the overall algorithm may not be suitable for real-time onboard usage.

\section{Problem Statement}\label{text:problem_statement}

We are motivated by the following scenario. Two mobile robots are moving through the indoor environment. Each robot is equipped with an RGB camera and a modern onboard GPU (e.g. NVidia Jetson) which are used for vision-based simultaneous localization and mapping. Each robot is performing vSLAM independently and at each time step outputs its pose and a 3D map of the environment. The task is to merge two individual maps into a single (global) map either at each time step or at the end of a mission.

%The 3D pointcloud maps are sliced along a selected plane and converted into occupancy grids for further 2D map-merging processing. Given the local maps and relative positions for each robot, we need to merge this maps into one global map (2D or 3D) and convert poses into global frame. 

Formally the problem can be defined as follows. Let $K$ be the current time step. At this time step two maps are given as the pointclouds:

\begin{equation}\label{eq:maps}
  M^1 = \{m^1_i\}, M^2 = \{m^2_j\}, i,j \in N, m^1_i,m^2_j \in R^3
\end{equation}

Similarly two poses are given:

\begin{equation}\label{eq:poses}
    P^1_k = (p^1_k,q^1_k), P^2_k = (p^2_k,q^2_k), k=1, 2, ..., K
\end{equation}

where $p^i_k = (x^i,y^i,z^i)_k$ represents a position of robot $i$ at time step $k$, and $q^i_k = (q^i_x, q^i_y,q^i_z,q^i_w)_k$ -- an orientation of the camera\footnote{We assume camera to be fixed firmly to the body of a robot}. These poses, represented as sequences correspond to trajectories of the robots: $T_1 = \{P^1_k~|~k=1,...,K\}$ and $T_2=\{P^2_k~|~k=1,...,K\}$

Map and trajectory merging are, formally, the functions:
\begin{equation}\label{eq:mapmerge}
    f_{mm}: 2^{R^3} \times 2^{R^3} \rightarrow 2^{R^3},  f_{tm}: 2^{R^7} \times 2^{R^7} \rightarrow 2^{R^7},
\end{equation}

where $M_e=f_{mm}(M^1, M^2)$ -- is the merged map and $T_e=f_{tm}(T^1, T^2)$ -- is the merged trajectory. 

In practice, these functions are often represented as $M_e = M_1 \cup A(M_2)$ (or $M_e = A(M_1) \cup M_2$), $T_e = T^1 \cup B(T^2)$ (or $T_e = B(T^1) \cup T^2$), where $A:R^3 \to R^3$ is usually represented (but no limited to) as affine transformation, and $B: R^7 \to R^7$ is an arbitrary transformation.

To measure the quality of the merged map/trajectory we assume that the ground truth map/trajectory is given, $M_{gt}$, $T_{gt}$, as well as the error functions, $L_M(M_e, M_{gt})$, $ L_T(T_e, T_{gt})$, that measure the error w.r.t. ground truth. 

\textbf{The merging problem} can now be represented as a tuple $\langle M^1, M^2, M_{gt}, \linebreak T^1, T^1, T_{gt}, L_M, L_T  \rangle$ and stated as:

\begin{equation}
    \begin{split}
        \mathrm{Find~}M_e, T_e \\
        \mathrm{s.t.} \\
        L_M(M_e, M_{gt}) \to min  \\
        L_T(T_e, T_{gt}) \to min
    \end{split}
\end{equation}

In this work we are interested mainly in map merging problem, thus we will omit the estimation of $T_e$ and the computation of $L_T$.

\begin{figure}
    \centering
    \includegraphics[width=1.0\textwidth]{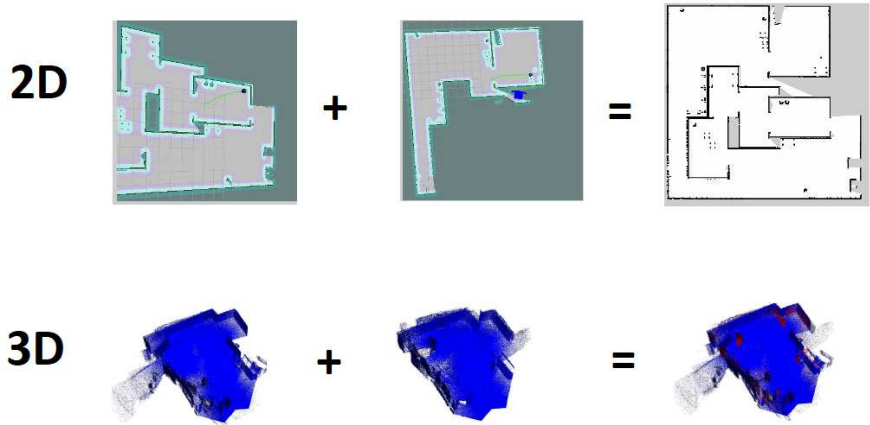}
    \caption{Map-merging in 2D and 3D.} 
    \label{fig:map_merge_examples}
\end{figure}

\begin{figure}[t]
    \centering
    \includegraphics[width=1.0\textwidth]{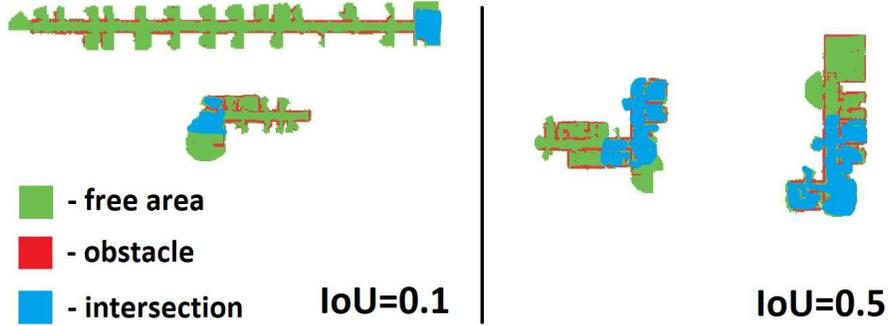}
    \caption{The IoU metric reflects the complexity of merging. Maps on the left overlap less than the maps on the right and that is captured by lower IoU (0.1 vs. 0.5). Thus it is harder to merge maps on the left.}
    \label{fig:map_complexity}
\end{figure}{}

\section{Methods and Metrics Overview}\label{text:methods_overview}

In this work we are interested both in 2D and 3D map-merging methods (see~\figurename~\ref{fig:map_merge_examples}. In order for the former to be applied to the maps constructed by a monocular vSLAM algorithm, these (pointcloud) maps should be first converted to 2D. We use grid representation of 2D maps:

%the 2D case as well. In order to use 2D methods, the input must be converted from 3D to 2D by either projecting pointlcloud on the plane (usually the ground plane, that is references to the start position of the robot) or by slicing:

%The occupancy grid is representing a projection or slice of 3D map:

\begin{equation}\label{eq:occ_grids}
    M^p = f_{pr}(M), M^p \in \{0,1,-1\}^{A \times B}
\end{equation}

where, $f_{pr}: R^3 \to R^2$ is a projection or slicing operation that results in a 2D grid composed of the $A \times B$ cells, each being either free -- 0, occupied -- 1, or unknown -- $-1$. %These grids can be visually represented as grayscale images -- see Fig.~\ref{fig:map_merge_examples}, middle and bottom rows.

To measure how difficult the merging task is we compute an intersection over union ($IoU \in (0;1]$) between the two individual maps:

\begin{equation}
    IoU = \frac{|M^1_{gt} \cap M^2_{gt}|}{|M^1_{gt} \cup M^2_{gt}|}
\end{equation}{}

Here, $M^i_{gt}$ stands for the ground-truth map (either 2D or 3D) for a robot $i$. The closer $IoU$ is to $0$ the less two map overlap. Thus it's harder to find correspondences between 2 maps, thus it is harder to merge the maps. On the contrary, if $IoU=1$ no computations are needed to merge as maps are equivalent. \figurename~\ref{fig:map_complexity} illustrates this concept.

To measure the error of merging we use the following error function:

\begin{equation}
    \begin{split}
        %G = min(M_e,M_{gt}) \\
        L_M=\frac{1}{G}\sum^{G}_{l=1}{\rho(m^l_e, m^{l}_{gt})} \\
        %d(m^e_l,m^{gt}_l) \to min
    \end{split}{}
\end{equation}{}

where $\rho$ is the proximity measure for an element of the merged map, $m_e$, and the corresponding element of the ground-truth map, $m_{gt}$, and $G$ is the number of elements (points) that compose the merged map. Intuitively this metric represent a proximity between two maps. In our experiments we used an implementation of this error-function which is a part of the CloudCompare software package\footnote{\url{http://www.cloudcompare.org/}}. It relies on the Euclidean distance as the proximity measure. To estimate the correspondence between $m_e$ and $m_{gt}$ it searches across all elements of $M_{gt}$ to find the one which is the closest to $m_e$. We will refer to this exact error-function as $L_{cc}$ further on.

We would also like to take into account the difficulty of merging when computing an error of the merge. This can be done by multiplying  $L{cc}$ by a scalar representing how difficult the merge is. In our case this measure of difficulty is $IoU$, so the scaled metrics is $wL_{CC} = IoU * L_{cc}$.

%However, in our previous work~\cite{inproceedings}, we found that this metric doesn't consider the vSLAM context of the input maps. So we developed the metric, that does map comparison based on positions and orientations from ground-truth and resultant map as a correspondenses to match the 3D or 2D pointclouds. We refer this metric as Absolute Mapping Error (AME).

%Also, we need to state of the merging process succeed with indicator $SR=\{0,1\}$ and find how complex the process of merging for particular maps. As complexity metric we define the coefficient as intersection over union (IoU):

Moreover we are also interested to assess the map merging outcome not only quantitatively but qualitatively as well. To this end we introduce the success metric. I.e. a human expert analyzes the merged map and reports whether the resultant map is a successful merge of the input or not.

%Summarizing everything written above, the evaluated algorithm needs to get a set of occupancy grids or poinclouds as inputs (2D and 3D maps accordingly) and output merged map. Then, we need to calculate proposed metrics in order to estimate the quality of the merge both for ground-truth maps and the maps, processed with modern CNN-based vSLAM algorithms.

\textbf{Algorithms} We chose 3 map-merging algorithms for our evaluation based on the following criteria: they have to be open-sourced and targeted for the application on real robots. First, the we chose map\_merge\_2d\footnote{\url{https://github.com/hrnr/m-explore}} and map\_merging\footnote{\url{https://github.com/emersonboyd/MultiSLAM}} algorithms (noted as \textbf{MM1} and \textbf{MM3} accordingly). These 2D algorithms require the occupancy grids as the input (so 3D pointclouds, obtained by vSLAM, need to be projected first as described above) and treat these grids as images to extract features from them which are further used to estimate the correspondences and merge the maps. The main difference between \textbf{MM1} and \textbf{MM3} is in feature detector method (SIFT for \textbf{MM1}, SURF for \textbf{MM3}).

Second, the map\_merge\_3d\footnote{\url{https://github.com/hrnr/map-merge}} (\textbf{MM2}) was chosen for 3D map-merging. This algorithm pre-proccesses the maps first to get the outliers removed with nearest neighbour outliter removal algorithm (if a point doesn't have neighbours in a certain area, it gets removed). Then, a 3D feature extraction algorithm detects the SIFT points or Harris corners on each map with corresponding descriptors. After this step, the features are compared to find the correspondences and transformation. If the match was successful the maps are merged.

\section{Experimental Evaluation and Results}\label{text:experiments}

\subsection{Setup}

\begin{figure}
    \centering
    \includegraphics[width=1.0\textwidth]{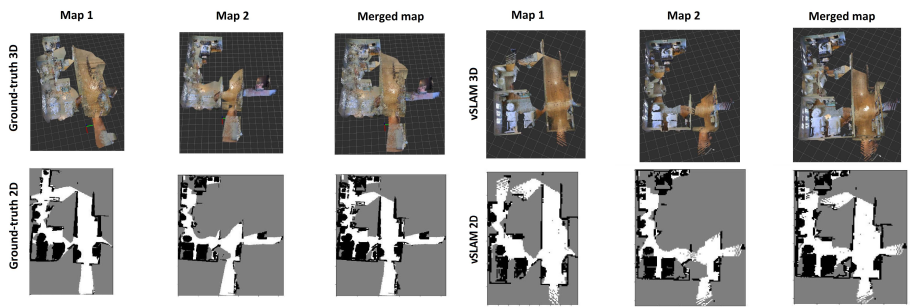}
    \caption{Part of the dataset used for the evaluation. On the left -- two separate maps, obtained by relying on ground-truth from the simulator, and a merged one. On the right -- the maps of the same environment constructed by a monocular vSLAM that did not have an access to ground-truth depths.}
    \label{fig:habitat}
\end{figure}{}

We used Robot Operating System~\cite{quigley2009ros} as the main framework. It  provides a set of ready-to-use algorithms for many robotics applications and also has tools for data visualization, recording, analysis etc. For modeling the navigation of robots through the indoor environments we chose Habitat simulator\footnote{\url{https://github.com/facebookresearch/habitat-sim}}. In this simulator we selected 6 scenes of Matterport 3D dataset~\cite{chang2017matterport3d}, that were built using motion capture systems. Habitat allows to simulate the movement of a robot (camera) through the environment and at each moment of time one is able to acquire the robot's pose in the world coordinate system, ground-truth depth image and an RGB image. Using this data we are able to reconstruct a map of the environment by direct re-projecting, i.e. without the need to run a SLAM algorithm. Such maps will be referred to as ground-truth ones.

\subsubsection{Individual trajectories} Overall we conducted 40 separate runs of a virtual robot through the simulated environment that were grouped into 20 overlapping pairs (see~\tablename~\ref{tab:env_setup}). During each run the set of RGB-images and depth maps as well as camera positions were recorded for further ground-truth map reconstruction. The dataset with corresponding rosbags is available at: \url{https://github.com/CnnDepth/matterport_overlapping_maps}.

\begin{table}[]
    \centering
    \begin{tabular}{c|c|c|c|c|c}
        Pair & \makecell{Trajectory 1 \\ length in m} & \makecell{Trajectory 2 \\ length in m} & IoU & Points 3D & Resolution 2D \\
    \hline
        1 & 21.7 & 20.3 & 0.480 & 405 288 & 399x403 \\
        2 & 32.9 & 20.8 & 0.295 & 493 070 & 448x509 \\
        3 & 13.7 & 15.9 & 0.619 & 130 162 & 234x399 \\
        4 & 19.0 & 22.9 & 0.449 & 415 810 & 463x419 \\
        5 & 18.4 & 13.8 & 0.437 & 175 628 & 383x243 \\
        6 & 16.2 & 21.2 & 0.402 & 193 740 & 444x219 \\
        7 & 4.1 & 7.8 & 0.479 & 121 247 & 239x236 \\
        8 & 13.9 & 11.0 & 0.457 & 163 418 & 329x250 \\
        9 & 9.4 & 4.3 & 0.302 & 88 337 & 211x245 \\
        10 & 9.8 & 10.0 & 0.399 & 189 478 & 250x310 \\
        11 & 21.5 & 11.2 & 0.399 & 240 041 & 330x379 \\
        12 & 16.3 & 17.4 & 0.327 & 397 487 & 528x407 \\
        13 & 22.2 & 21.0 & 0.535 & 266 472 & 329x425 \\
        14 & 13.9 & 15.8 & 0.420 & 214 593 & 327x484 \\
        15 & 12.6 & 21.1 & 0.202 & 277 755 & 397x314 \\
        16 & 21.6 & 17.8 & 0.266 & 267 133 & 400x284 \\
        17 & 10.8 & 6.5 & 0.388 & 164 773 & 349x238 \\
        18 & 13.7 & 12.3 & 0.512 & 156 416 & 255x326 \\
        19 & 15.6 & 14.1 & 0.417 & 183 705 & 228x319 \\
        20 & 18.9 & 19.9 & 0.283 & 251 173 & 293x465 \\
    \end{tabular}
    \caption{Trajectories and maps used in experimental evaluation.}
    \label{tab:env_setup}
\end{table}

The lengths of the obtained trajectories varied from 4.1 m to 32.9 m. The intersection was between 20\% and 62\%. When constructing the pointclods we used 5cm as a resolution. With this resolution the number of points that constitute the maps varied from 88 337 to 493 070. The corresponding 2D projections (grids) varied in sized as shown in the last column of~\tablename~\ref{tab:env_setup}.

The experimental hardware/software setup was as follows:

\begin{itemize}
    \item CPU - Intel Core i5, 6-core
    \item GPU - GeForce RTX 2060
    \item RAM - 32 Gb
    \item OS: Ubuntu 18.04
    \item ROS version: Melodic
\end{itemize}

\subsection{Results}

\subsubsection{Reconstructing individual maps}

Besides ground-truth maps from the simulator we also used the following SLAM-constructed maps. First, we ran two RGB-D SLAM algorithms on the data from the simulator to construct the individual maps for further merging. These algorithms are: RTAB-Map\footnote{\url{https://github.com/introlab/rtabmap}}~\cite{labbe2019rtab} and RGBDSLAM\_v2\footnote{\url{https://github.com/felixendres/rgbdslam\_v2}}~\cite{endres20133}. Next, we constructed individual maps by relying only on RGB video, i.e. we infer depth data from RGB by a convolutional neural network \cite{inproceedings}~\footnote{\url{https://github.com/CnnDepth/tx2\_fcnn\_node}} and used this data in RTAB-Map and RGBDSLAM\_v2 to construct the final maps.

%The results of vSLAM output maps for RTAB-Map and RGBSLAM\_v2 with ground-truth depth maps are shown in 

\tablename~\ref{tab:slam_results_gt} shows the accuracy of the maps constructed by RTAB-Map and RGBDSLAM\_v2 relying on RGB-D data from simulator, i.e. images and ground-truth depths for each pixel on that images. As one can see RGBDSLAM\_v2 maps are less accurate thus it is expected that these maps should be harder to merge compared to the ones produces by RTAB-Map.

\begin{table}[]
    \centering
    \begin{tabular}{c|c|c|c|c|c|c}
             & \multicolumn{2}{c|}{RTAB-Map+gt-depth} & \multicolumn{2}{c|}{RGBDSLAM\_v2+gt-depth}\\
    \hline 
        Pair & Map1 $L_{CC}$ & Map2 $L_{CC}$ & Map1 $L_{CC}$ & Map2 $L_{CC}$  \\
    \hline
        1 & 0.142 & 0.081 & 0.277 & 0.090 \\
        2 & 0.377 & 0.137 & 0.242 & 0.249 \\
        3 & 0.032 & 0.155 & 0.080 & 1.188 \\
        4 & 0.310 & 0.081 & 1.115 & 0.123 \\
        5 & 0.054 & 0.122 & 0.138 & 0.040 \\
        6 & 0.170 & 0.120 & 0.289 & 0.564 \\
        7 & 0.071 & 0.048 & 0.052 & 0.049 \\
        8 & 0.105 & 0.078 & 0.119 & 0.284 \\
        9 & 0.048 & 0.095 & 0.053 & 0.030 \\
        10 & 0.062 & 0.080 & 0.040 & 0.258 \\
        11 & 0.238 & 0.052 & 0.291 & 0.053 \\
        12 & 0.068 & 0.111 & 0.954 & 0.373 \\
        13 & 0.061 & 0.097 & 8.425 & 0.285 \\
        14 & 0.071 & 0.089 & 0.408 & 0.370 \\
        15 & 0.112 & 0.074 & 0.201 & 0.312 \\
        16 & 0.246 & 0.086 & 0.380 & 1.042 \\
        17 & 0.035 & 0.050 & 0.226 & 0.116 \\
        18 & 0.044 & 0.117 & 0.097 & 0.307 \\
        19 & 0.091 & 0.179 & 0.094 & 0.267 \\
        20 & 0.086 & 0.088 & 0.424 & 0.615 \\

    \end{tabular}
    \caption{Accuracy of the RTAB-MAP and RGBDSLAM\_v2 maps that were constructed using the ground-truth depths from the simulator.}
    \label{tab:slam_results_gt}
\end{table}

\begin{table}[]
    \centering
    %\resizebox{0.6\columnwidth }{!}{
    \begin{tabular}{c|c|c|c|c|c|c}
        Pair & Map1 $L_{CC}$ & Map2 $L_{CC}$ & Map1 $L_{CC}$ (scaled) & Map2 $L_{CC}$ (scaled) \\
    \hline
        1 & 0.588 & 0.483 & 0.253 & 0.321\\
        2 & 2.598 & 0.490 & 1.448 & 0.180\\
        3 & 1.677 & 1.067 & 0.196 & 0.217\\
        4 & 0.453 & 0.385 & 0.313 & 0.385\\
        5 & 1.696 & 1.272 & 0.263 & 0.242\\
        6 & 1.389 & 2.047 & 0.359 & 0.227\\
        7 & 0.694 & 0.653 & 0.415 & 0.256\\
        8 & 1.399 & 1.263 & 0.225 & 0.359\\
        9 & 0.818 & 0.848 & 0.336 & 0.153\\
        10 & 0.420 & 0.746 & 0.391 & 0.356\\
        11 & 1.078 & 0.428 & 0.233 & 0.333\\
        12 & 0.315 & 0.429 & 0.268 & 0.358\\
        13 & 0.468 & 0.454 & 0.304 & 0.309\\
        14 & 0.336 & 0.300 & 0.278 & 0.233\\
        15 & 0.409 & 1.123 & 0.305 & 0.228\\
        16 & 0.615 & 1.294 & 0.300 & 0.211\\
        17 & 1.102 & 0.546 & 0.192 & 0.214\\
        18 & 1.361 & 1.399 & 0.205 & 0.213\\
        19 & 2.164 & 1.951 & 0.217 & 0.168\\
        20 & 2.510 & 1.798 & 0.182 & 0.407\\

    \end{tabular}
    %}
    \caption{Accuracy of the maps constructed by RTAB-MAP using CNN inferred depths.}
    \label{tab:slam_results_fcnn}
\end{table}

\tablename~\ref{tab:slam_results_fcnn} shows the accuracy of the maps produced by RTAB-Map algorithm relying on CNN reconstructed depths. In this case, we found that FCNN algorithm reconstructs the depth with constant scale error for each map\footnote{We believe the main reason for this was that the training of CNN had been conducted on a different dataset }. Thus we automatically adjusted this scale factor for each map in order to minimize the mismatch from the ground-truth. Unfortunately RGBDSLAM\_v2 algorithm was not able to construct a single individual map relying on CNN-inferred depths, thus \tablename~\ref{tab:slam_results_fcnn} does not contain information on  RGBDSLAM\_v2 + CNN.

%We refer this error as $L_{CC}$ scaled. The original $L_{CC}$ is 2-20 times worse compared to ground-truth (both RTAB-Map and RGBD-SLAM\_v2) produced maps, however, the scaled variants sometimes are even more accurate. 

\begin{table}
    \centering
    \resizebox{\columnwidth }{!}{%
    \begin{tabular}{c|c|c|c|c|c|c|c|c|c}
        Pair & MM1 & MM2  & MM3  & MM1 $L_{CC}$ & MM2 $L_{CC}$ & MM3 $L_{CC}$ & MM1 $wL_{CC}$ & MM2 $wL_{CC}$ & MM3 $wL_{CC}$ \\
    \hline
%        1 & - & - & - & x & x & x & x & x & x\\
%        2 & - & - & - & x & x & x & x & x & x\\
        3 & + & + & + & 0.080 & 0.168 & 0.155 & 0.050 & 0.104 & 0.096\\
%        4 & - & - & - & x & x & x & x & x & x\\
        5 & + & + & + & 0.464 & 0.127 & 0.139 & 0.203 & 0.055 & 0.061\\
        6 & - & + & + & x & 0.104 & 0.108 & x & 0.042 & 0.043\\
        7 & + & + & + & 0.063 & 0.086 & 0.076 & 0.030 & 0.041 & 0.036\\
%        8 & - & - & - & x & x & x & x & x & x\\
        9 & - & + & - & x & 0.127 & x & x & 0.038 & x\\
        10 & + & - & - & 0.124 & x & x & 0.049 & x & x\\
        11 & + & - & - & 0.159 & x & x & 0.063 & x & x\\
        12 & - & + & - & x & 0.110 & x & x & 0.036 & x\\
        13 & + & + & + & 0.148 & 0.119 & 0.078 & 0.079 & 0.064 & 0.042\\
        14 & + & + & + & 0.069 & 0.118 & 0.229 & 0.029 & 0.050 & 0.096\\
%        15 & - & - & - & x & x & x & x & x & x\\
        16 & - & + & + & x & 0.096 & 0.065 & x & 0.026 & 0.017\\
%        17 & - & - & - & x & x & x & x & x & x\\
        18 & - & + & + & x & 0.117 & 0.103 & x & 0.060 & 0.053\\
        19 & - & + & - & x & 0.149 & x & x & 0.062 & x\\
        20 & + & - & + & 0.080 & x & 0.261 & 0.023 & x & 0.074\\

    \end{tabular}
    }
    \caption{Merging results for RTAB-MAP + ground truth depths.}
    \label{tab:merge_results_habitat}
\end{table}

\subsubsection{Map-merging}

The results of running different map-merging algorithms on the individual maps constructed by RTAB-Map provided with ground-truth depths are shown in~\tablename~\ref{tab:merge_results_habitat}. To estimate $L_{CC}$ error of merged 2D maps, we used projection provided by RTAB-Map (in \textit{/proj\_map} topic). Successful merges are denoted with $+$ sign. Accuracy metric, $L_{cc}$, is shown as well\footnote{Please note, that in case of \textbf{MM1} we have to find a transform between the merged map and the ground-truth map manually in order to compute $L_{cc}$, as the algorithm does not provide this transform. We used the transform that minimized  $L_{cc}$.}.

MM2 algorithm was able to merge 11 pairs of maps, compared to 8 for MM1 and 9 for MM3. The lowest error, $L_{cc}$, was obtained with MM2 algorithm on pair 14, however, the close results were obtained with MM1. None of the algorithms were able to merge the most challenging map pair (pair 15 with IoU=0.202), however the MM2 algorithm was able to merge pair 16 with IoU=0.266 and achieved $wL_{CC}=0.026$.

The results of running map-merging methods on the individual maps constructed by RGBDSLAM\_v2 provided with ground-truth depths are shown in~\tablename~\ref{tab:merge_results_rgbdslam}. Recall, that for RGBDSLAM\_v2 the $L_{CC}$ of the individual maps was poor thus only 6 pairs of maps were successfully merged and $L_{CC}$ of the merged maps varied from 0.106 to 0.448 (this is 2-5 times worse compared to RTAB-Map + ground truth). We got 4 successful merges from MM1 algorithm, 5 merges be the MM2 and 1 successful merge from MM3 algorithm.%In general, it is evident that additional algorithms and methods need to be developed in order to successfully merge CNN-based vSLAM maps.

\begin{table}
    \centering
    \resizebox{\columnwidth }{!}{
    \begin{tabular}{c|c|c|c|c|c|c|c|c|c}
        Pair & MM1 & MM2 & MM3 & MM1 $L_{CC}$ & MM2 $L_{CC}$ & MM3 $L_{CC}$ & MM1 $wL_{CC}$ & MM2 $wL_{CC}$ & MM3 $wL_{CC}$\\
    \hline
%        1 & - & x & x \\
%        2 & - & x & x \\
%        3 & - & x & x \\
%        4 & - & x & x \\
         5 & + & - & + & 0.072 & x & 0.147 & 0.028 & x & 0.057\\
%        6 & - & x & x \\
        7 & + & + & - & 0.056 & 0.106 & x & 0.016 & 0.031 & x \\
%        8 & - & x & x \\
        9 & + & + & - & 0.393 & 0.248 & x & 0.097 & 0.061 & x \\
%        10 & - & x & x \\
        11 & + & + & - & 0.255 & 0.448 & x & 0.054 & 0.095 & x \\
%        12 & - & x & x \\
%        13 & - & x & x \\
%        14 & - & x & x \\
%        15 & - & x & x \\
%        16 & - & x & x \\
        17 & - & + & - & x & 0.303 & x & x & 0.117 & x \\
        18 & - & + & - & x & 0.203 & x & x & 0.097 & x \\
%        19 & - & x & x \\
%        20 & - & x & x \\
    \end{tabular}
    }
    \caption{Merging results for RGBDSLAM\_V2 algorithm + ground-truth depths.}
    \label{tab:merge_results_rgbdslam}
\end{table}

We also tried to merge maps constructed by RTAB-Map relying on CNN-estimated depths. Each of MM1, MM2 and MM3 algorithms was able to merge only 2 pairs of maps. Merging results are shown in~\tablename~\ref{tab:merge_results_fcnn}. The lowest $L_{CC}$ error was 0.369, the highest $L_{CC}$ error was 2.209. The error of maps merged by MM1 and MM2 algorithms was less than the error of corresponding individual maps, and the error of maps merged by MM3 was more than error of individual maps. However MM3 was able to merge challenging pair 16 (IoU = 0.266), whereas MM1 and MM2 merged only pairs with IoU more than 0.4.

\begin{table}[]
    \centering
    \resizebox{\columnwidth }{!}{%
    \begin{tabular}{c|c|c|c|c|c|c|c|c|c}
        Pair & MM1 & MM2 & MM3 & MM1 $L_{CC}$ & MM2 $L_{CC}$ & MM3 $L_{CC}$ & MM1 $wL_{CC}$ & MM2 $wL_{CC}$ & MM3 $wL_{CC}$ \\
    \hline
        %1 & - & - & - & x & x & x & x & x & x\\
        %2 & - & - & - & x & x & x & x & x & x\\
        3 & + & - & + & 0.821 & 0.982 & x & 0.508 & 0.608 & x\\
        4 & + & - & - & 0.369 & x & x & 0.166 & x & x\\
        %5 & - & - & - & x & x & x & x & x & x\\
        6 & - & + & + & x & 1.611 & 2.209 & x & 0.648 & 0.888\\
        %7 & - & - & - & x & x & x & x & x & x\\
        %8 & - & - & - & x & x & x & x & x & x\\
        %9 & - & - & - & x & x & x & x & x & x\\
        %10 & - & - & - & x & x & x & x & x & x\\
        %11 & - & - & - & x & x & x & x & x & x\\
        %12 & - & - & - & x & x & x & x & x & x\\
        %13 & - & - & - & x & x & x & x & x & x\\
        %14 & - & - & - & x & x & x & x & x & x\\
        %15 & - & - & - & x & x & x & x & x & x\\
        16 & - & + & - & x & x & 1.391 & x & x & 0.370\\
        %17 & - & - & - & x & x & x & x & x & x\\
        %18 & - & - & - & x & x & x & x & x & x\\
        %19 & - & - & - & x & x & x & x & x & x\\
        %20 & - & - & - & x & x & x & x & x & x\\

    \end{tabular}
    }
    \caption{Merging results with RTABMAP + CNN-inferred depths.}
    \label{tab:merge_results_fcnn}
\end{table}

Overall, the obtained results show that map merging is a hard-to-accomplish task even for maps that were constructed by vSLAM algorithms that have access to ground-truth depths. In case depth data is not available during SLAM but rather has to be inferred with CNN the quality of the individual maps degrade to the extent that map-merging becomes extremely hard.

%First we tried to merge maps builded by RGBDSLAMv2 with FCNN depths. The results were unsatisfactory - RGBDSLAMv2 failed to build the map in all 40 cases. So there were no successful merges with RGBDSLAMv2 + FCNN. Next we tried to merge the maps obtained with RTAB-Map + FCNN. We got 2 successful merges from each of MM1, MM2, and MM3 algorithms. $L_{CC}$ error of merged maps varied from 0.369 to 2.209 whereas $L_{CC}$ error of individual maps built by RTAB-MAP + FCNN varied from 0.300 to 2.598.

%This is due to the jitters and overall low quality (in comparison to ground-truth) of the maps, and due to feature based algorithms, used during map-merging (no exact patterns are found during correspondences search).

%\begin{table}[]
%    \centering
%    \begin{tabular}{c|c|c|c|c|c|c}
%        Pair & First CC & Second CC \\
%    \hline
%        12 & 0.206 & 0.192 \\
%        13 & 3.115 & 0.151 \\
%        14 & 0.085 & 0.097 \\
%        15 & 0.225 & 0.475 \\
%        16 & 0.273 & 0.661 \\
%        17 & 0.158 & 0.058 \\
%        18 & 0.096 & 0.314 \\
%        19 & 0.081 & 0.246 \\
%        20 & 0.493 & 0.812 \\

%    \end{tabular}
%    \caption{CloudCompare metrics values for RGBDSLAM\_V2 maps}
%    \label{tab:metrics_rgbdslam}
%\end{table}

\section{Conclusion}\label{text:conclusion}

In this paper, we have considered a map-merging problem when the source maps come from visual-based SLAM algorithms. Metric that reflects the hardness of merging task as well as the one reflecting the quality of the merge was described. To evaluate different map-merging approaches a large dataset was created via the Habitat simulator environment. Using this simulator allowed us to get access to ground-truth map models that are very hard to construct in real-world experiments. Relying on this ground-truth data we evaluated several map-merging algorithms (both 2D and 3D) paired with two different vSLAM algorithms. The results of empirical evaluation provide a clear evidence that map-merging of vSLAM-constructed maps is a non-trivial problem that is lacking general solution yet, i.e. numerous problem instances remain unsolved by state-of-the-art merging algorithms. This clearly provides avenues for future research, especially for 3D map-merging.

\subsubsection{Acknowledgements} 

This work was supported by Russian Science Foundation project \#16-11-0048 (CNN-based depth reconstruction, RTAB-MAP+FCNN implementation, experimental evaluation of map-merging algorithms) and by the ``RUDN University Program 5-100'' (data preparation).

\bibliographystyle{splncs04}
\bibliography{samplepaper}
%Author, F.: Article title. Journal \textbf{2}(5), 99--110 (2016)

%Author, F., Author, S.: Title of a proceedings paper. In: Editor,
%F., Editor, S. (eds.) CONFERENCE 2016, LNCS, vol. 9999, pp. 1--13.
%Springer, Heidelberg (2016). \doi{10.10007/1234567890}

%Author, F., Author, S., Author, T.: Book title. 2nd edn. Publisher,

%Author, A.-B.: Contribution title. In: 9th International Proceedings
%on Proceedings, pp. 1--2. Publisher, Location (2010)

%LNCS Homepage, \url{http://www.springer.com/lncs}. Last accessed 4
%\end{thebibliography}
\end{document}